\documentclass[runningheads]{llncs}
\pdfoutput=1
\usepackage{times}
\usepackage{epsfig}
\usepackage{graphicx}
\usepackage{amsmath}
\usepackage{amssymb}

\usepackage{booktabs}
\usepackage{subcaption}
\usepackage{float}
\usepackage{multirow}
\usepackage{comment}
\usepackage{csquotes}
\usepackage{multicol}
\usepackage{tabularx}
\usepackage{xcolor}
\usepackage{wrapfig}
\usepackage{afterpage}
\usepackage[toc,page]{appendix}

\newcolumntype{M}[1]{>{\centering\arraybackslash}m{#1}}
\newcommand*{\ff}[1]{\textcolor{blue}{#1}}

\begin{document}
\title{Episode Adaptive Embedding Networks for\\ Few-shot Learning}

%
\author{Fangbing Liu\inst{1} \and
Qing Wang\inst{1}}
\institute{Australian National University\\
\email{fangbing.liu@anu.edu.au},\hspace{0.3cm}\email{qing.wang@anu.edu.au}}

%
\maketitle              
\begin{abstract}
Few-shot learning aims to learn a classifier using a few labelled instances for each class. Metric-learning approaches for few-shot learning embed instances into a high-dimensional space and conduct classification based on distances among instance embeddings. However, such instance embeddings are usually shared across all episodes and thus lack the discriminative power to generalize classifiers according to episode-specific features. In this paper, we propose a novel approach, namely \emph{Episode Adaptive Embedding Network} (EAEN), to learn episode-specific embeddings of instances. By leveraging the probability distributions of all instances in an episode at each channel-pixel embedding dimension, EAEN can not only alleviate the overfitting issue encountered in few-shot learning tasks, but also capture discriminative features specific to an episode. To empirically verify the effectiveness and robustness of EAEN, we have conducted extensive experiments on three widely used benchmark datasets,  under various combinations of different generic embedding backbones and different classifiers. The results show that EAEN significantly improves classification accuracy about $10\%$ to $20\%$ in different settings over the state-of-the-art methods.

\end{abstract}
\section{Introduction}
Few-shot learning has attracted attention recently due to its potential to bridge the gap between the cognition ability of humans and the generalization ability of machine learning models \cite{matchingnets,maml,propagation,infinite_pro}. At its core, few-shot learning aims to learn a classifier using a few labelled instances for each class. This however poses significant challenges to traditional machine learning algorithms which are designed to learn from a large amount of labelled instances. They easily overfit when trained on a small training set, and thus fail to generalize to new classes.

Driven by a simple learning principle: ``test and train conditions must match", \emph{episode training} was proposed to deal with the few-shot learning problem \cite{matchingnets}. In the episode training setting, each episode contains only a few labelled instances per class (i.e., \emph{support set}) and a number of unlabelled instances (i.e., \emph{query set}) whose classes are to be predicted. Thus, an episode mimics a classification task in few-shot learning scenarios, and a learning model can be trained by conducting a series of classification tasks moving from episode to episode. As reported in \cite{matchingnets,few_gnn}, compared with traditional supervised training in which labelled instances are from one classification task, episode training leads to a better generalization ability on small training data.

\begin{wrapfigure}{r}{0.6\textwidth}
\vspace{-0.9cm}
 \centering
    \includegraphics[width=0.6\columnwidth]{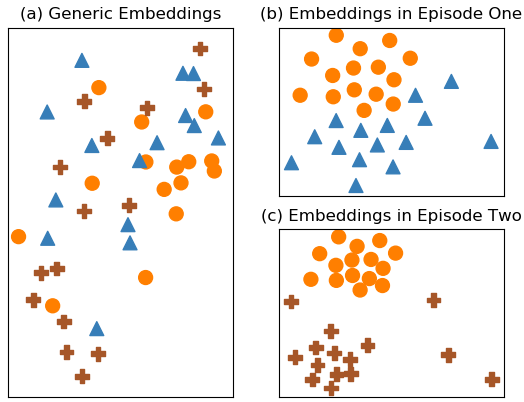}
    \caption{Instance embeddings (a) in a generic embedding space, and (b)-(c) in an episode-specific embedding space.}
    \label{fig_emb}\vspace{-0.6cm}
\end{wrapfigure}

Inspired by \cite{matchingnets}, episode training has been adopted in many later studies for few-shot learning \cite{baseline++,edgelabel,maml}. 
One promising research stream focuses on developing metric-learning-based approaches with episode training \cite{matchingnets,prototypical,infinite_pro,relationnets}. The key idea is to map instances into a high-dimensional embedding space such that their embeddings capture  discriminative features for classification. Then, distances between instance embeddings are measured, and unlabelled instances in an episode are classified according to their distances with labelled instances. Although achieving reasonably good performance, most approaches do not consider features specific to classification tasks when embedding instances, i.e., episode-specific features. 
For example, instances of three classes ``dog" (circle), ``cat" (cross) and ``wolf" (triangle) can be mapped into generic embeddings shown in Figure \ref{fig_emb}(a), without considering their episode-specific features. However, it is hard to classify these instances based on their generic embeddings. By embedding instances into an episode-specific embedding space that capture episode-specific features, such as features distinguishing ``dog" from ``wolf", or ``dog" from ``cat", as shown in Figure \ref{fig_emb}(b)-(c), it is easier to learn classification boundaries within an episode.

Recently, some works \cite{set2set,task_aware,task_related,han2018meta} began to explore instance embeddings specific to classification tasks in few-shot learning. They have generally followed two directions: (a) tailoring the embeddings of support instances (i.e., instances in a support set) by learning their inter-class discriminative features within an episode \cite{set2set,task_aware,task_related}; (b) adjusting the embeddings of query instances (i.e., instances in a query set) according to their characteristics \cite{han2018meta}. For example, support instances were used to refine their generic embeddings via a set-to-set function in \cite{set2set}. 
A task-aware feature embedding network was introduced in \cite{task_aware} to adjust instance embeddings for specific tasks in a meta-learning framework. Nevertheless, none of these methods have fully captured episode-specific features into instance embeddings. They focused on extracting features specific to classes and to instances, whereas neglecting to account for features that align query instances with support instances in a specific episode. Instance embeddings thus lack the discriminative ability to generalize classifiers across episodes with new classes. Moreover, since only a few instances are available in a support set in few-shot learning, the low-data problem also hinders classification performance of these methods.

To circumvent these limitations, we propose \emph{Episode Adaptive Embedding Networks} (EAENs), which leverage the probability distributions of \emph{all instances} in an episode, including instances from both a support set and a query set, to extract representative episode-specific features. Particularly, EAENs consider the probability distributions of all instances in an episode at \emph{each channel-pixel embedding dimension}. This leads to an effective adaptation that transforms generic embeddings into episode-specific embeddings for improved generalisation. Thus, unlike prior works, EAENs have two distinct advantages. First, it alleviates the overfitting issue since it learns based on embeddings of both support and query instances, in contrast to just a few support instances per class used in existing works.
Second, it captures features that align query instances with support instances in each specific episode into embeddings. This is important for improving classification performance because metric-learning approaches for few-shot learning mostly rely on measuring distances among instance embeddings. 
In summary, our main contributions are as follows:
\begin{itemize}
    \item We propose a novel approach (EAENs) for few-shot learning, which maps instances into an episode-specific embedding space, capturing episode-specific features. 
    
    \item We derive formulae to exhibit the probability distributions of all instances in an episode with respect to each channel-pixel embedding dimension. This improves the generalization ability of classifiers.
    \item We conduct experiments to verify the effectiveness and robustness of our approach. Compared with the state-of-the-art models, our approach achieves about $20\%$ accuracy improvement in $5$-way $1$-shot and about $10\%$ improvement in $5$-way $5$-shot on both miniImageNet and tieredImageNet datasets, as well as competitive performance on CIFAR-FS dataset.
\end{itemize}

\section{Related Work}
Few-shot learning has been extensively studied in
recent years \cite{matchingnets,baseline++}. Our work in this paper is broadly related to three streams of research in few-shot learning.

\medskip
\noindent\textbf{Metric-learning approaches.~}The key idea behind metric-learning approaches is to learn instance  embeddings such that discriminative features of instances can be captured by their embeddings in a high-dimensional space \cite{matchingnets,prototypical,infinite_pro,relationnets,task_related,covariance}. Then, a distance-based classifier is employed to classify instances based on distances between instances in their embedding space. To avoid the overfitting problem in few-shot learning, these approaches often use simple non-parametric classifiers, such as nearest neighbor classifiers \cite{matchingnets,prototypical,infinite_pro}. Distances between instance embeddings are typically measured by simple L$1$ and cosine distances \cite{matchingnets}. A recent work proposed to learn such a distance metric for comparing instances within episodes \cite{relationnets}. 

\medskip
\noindent\textbf{Meta-learning approaches.~}Lots of meta-learning approaches have been proposed for few-shot learning tasks \cite{fine-grain,tadam,train_maml,reptile}. These approaches aim to minimize generalization error across different tasks and expect that a classifier performs well on unseen tasks \cite{maml,tadam,train_maml,reptile}.  
However, they mostly only learn generic embeddings that are same for all tasks. Some recent works have studied task-related embeddings \cite{opt,task_aware}. Since only a few labelled instances are available for each unseen class in a target task, learning discriminative task-related embeddings is hard and these works implicitly relied on the alignment of data distributions between seen and unseen classes. Several works also used data hallucination methods to synthesize instances to help classification \cite{augment1,augment2}.  



\medskip
\noindent\textbf{Transductive approaches.~}Depending on whether instances in a query set (i.e., query instances) are taken into account when designing a learning model, approaches for few-shot learning can be categorized as being transductive and non-transductive. Several works used query instances and their structure in episodes to conduct a classification task in a transductive way \cite{propagation,few_gnn,edgelabel,dpgn}. A label propagation method was proposed in \cite{propagation}, where label information was propagated from instances in a support set to instances in a query set. Graph neural networks were employed to diffuse information from neighbor instances for better embeddings \cite{few_gnn}. Assuming that all instances are fully connected with each other, \cite{edgelabel} proposed an iterative edge-labeling algorithm to predict edge labels, i.e., whether two instances connected by an edge belong to the same class.

\section{Episode Adaptive Embedding Networks}
We formulate the few-shot classification problem using \emph{episode training} \cite{matchingnets}. Let $\mathcal{D}$ be a set of classes which consists of two disjoint subsets $\mathcal{D}_{train}$ and $\mathcal{D}_{test}$. In a $N$-way $K$-shot setting, we randomly sample $N$ classes from $\mathcal{D}_{train}$, and then randomly sample $K$ instances for each class to form a \emph{support set} $S=\{(\mathbf{x}_i, y_i)\}_{i=1}^{N\times K}$ and $T$ instances for each class to form a \emph{query set} ${Q}=\{(\mathbf{x}_j, y_j)\}_{j=1}^{N\times T}$ in an episode, where $y_i$ is the class of an instance $\mathbf{x}_i$.
A classifier is trained to predict the classes of instances in the query set ${Q}$, which are compared with their true classes to calculate losses in training.

\begin{figure*}
\vspace{-0.5cm}
    \centering
    \includegraphics[width=0.9\textwidth]{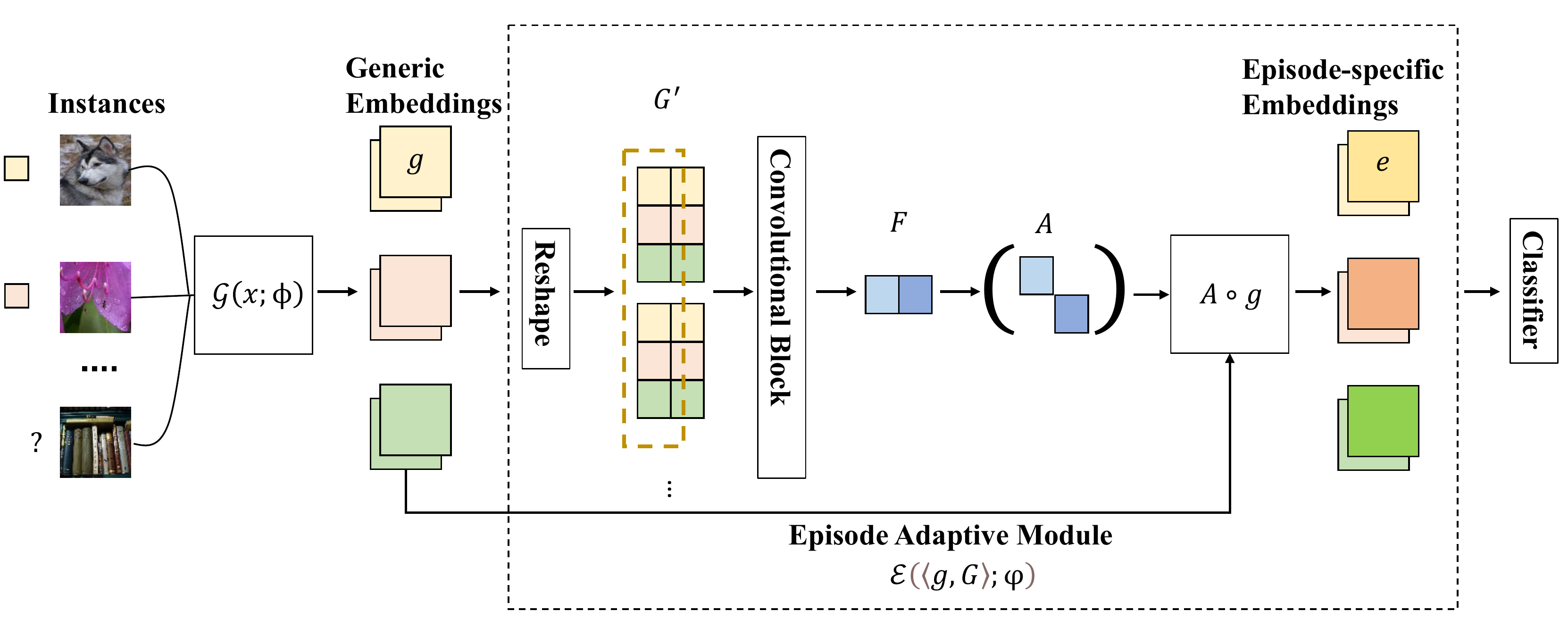}
    \caption{The framework of Episode Adaptive Embedding Networks.}
    \label{fig_model}\vspace{-1cm}
\end{figure*}

We propose \emph{Episode Adaptive Embedding  Networks} (EAENs) for few-shot classification, which consists of three components: a generic embedding module, an episode adaptive module and a classifier, as illustrated in Figure \ref{fig_model}. 





\subsection{Generic Embedding Module}

We define a generic embedding module $\mathcal{G}(\mathbf{x};\phi)$ to be a convolutional block $\mathcal{G}$ with learnable parameters $\mathbf{\phi}$. Given an instance $\mathbf{x}\in {\mathbb{R}^{w\times h \times c}}$ where $w$ and $h$ are the width and height of an instance, respectively, and $c$ refers to the number of its channels, a generic embedding module $\mathcal{G}$ takes $\mathbf{x}$ as input and embeds it to a three-dimensional tensor $\mathbf{g}\in{\mathbb{R}^{w'\times h'\times c'}}$, where $w'$, $h'$ and $c'$ represent the width, height, and number of channels of instance embeddings in a generic embedding space, respectively.

Let $E=(S,Q)$ denote an episode consisting of a support set $S$ and a query set $Q$. By applying a generic embedding module $\mathcal{G}(\mathbf{x};\phi)$ on $E$, we obtain the generic embeddings of all instances in $S$ and $Q$. For simplicity, we use a \emph{generic embedding matrix} $\mathbf{G}\in \mathbb{R}^{m\times n}$ to represent the generic embeddings of all instances from the episode $E$, where $m=w'\times h'\times c'$ and $n=N\times (K+T)$.

An instance may appear in one or more episodes. However, given an instance $\mathbf{x}$, the generic embeddings of $\mathbf{x}$ are always same for all episodes. In other words, a generic embedding module $\mathcal{G}(\mathbf{x};\phi)$ embeds instances into a generic embedding space without taking into account episodes to which instances belong.

\subsection{Episode Adaptive Module}
\vspace{-0.2cm}
An episode adaptive module $\mathcal{E}(\langle\mathbf{g},\mathbf{G}\rangle;\mathbf{\varphi})$ is defined as a neural network $\mathcal{E}$ with parameters $\varphi$. It takes $\langle\mathbf{g},\mathbf{G}\rangle$ as input, where $\mathbf{g}$ is the generic embeddings of an instance and $\mathbf{G}$ is the generic embedding matrix of an episode $E$ that the instance belongs to, and produces an episode-specific embeddings for the instance w.r.t. the episode $E$.

Specifically, for each episode $E$, we first reshape its generic embedding matrix $\mathbf{G}$, which contains the generic embeddings of all instances from $E$, into a three-dimensional tensor $\mathbf{G}'\in {\mathbb{R}^{m \times n\times 1}}$. Then, we feed $\mathbf{G}'$ as input to three convolutional layers in order to extract episode-specific features from generic embeddings based on a channel-pixel adaptive mechanism. This process yields episode-specific adaptive vector, each of its element corresponds to a channel-pixel value, to transform instance embeddings from a generic embedding space to an episode-specific embedding space.


Let $\mathbf{G}'(uvk,:,:)\in{\mathbb{R}^{n\times 1}}$ denote a matrix of instance embeddings at a fixed channel-pixel $uvk$, i.e., generic embeddings at the location $(u,v)$ of the $k$-th channel in $\mathbf{G}'$, where $u\in{[0,w')}$, $v\in{[0,h')}$ and $k\in{[0,c')}$. Then, we extract episode-specific features from $\mathbf{G}'(uvk,:,:)$ with a convolutional block which successively applying three convolutional layers with decreasing numbers of kernels (e.g., 64 kernels for the first layer, 32 kernels for the second layer, and 1 kernel for the third layer):

  \begin{equation}
    \mathbf{P}(uvk,:,i)=\sigma(\mathbf{W}_i^{p}\circ \mathbf{G}'(uvk,:,:))\text{\hspace*{1cm}} \text{for \hspace*{0.1cm}} i=1,\dots, d;
 \end{equation}
 \begin{equation}
     \mathbf{Z}(uvk,:,j)=\sigma(\mathbf{W}_j^{z}\circ \mathbf{P}(uvk,:,:))\text{\hspace*{1.12cm}} \text{for \hspace*{0.1cm}} j=1,\dots, f;
\end{equation}
\begin{equation}
    \mathbf{F}(uvk,:,:)=\sigma(\mathbf{W}^{a}\circ \mathbf{Z}(uvk,:,:)). \text{\hspace*{3.6cm}}
\end{equation}
\vspace{0cm}

\noindent where $\mathbf{W}_i^{p}\in{\mathbb{R}^{1\times n}}$, $\mathbf{W}_j^{z}\in{\mathbb{R}^{1\times d}}$ and $\mathbf{W}^{a}\in{\mathbb{R}^{1\times f}}$ are the parameters of the $i$-th kernel of the first convolutional layer, the $j$-th kernel of the second convolutional layer and the only kernel of the third convolutional layer, respectively, $\circ$ denotes a matrix multiplication, and $\sigma$ is a non-linear activation function. After extracting episode-specific features from every channel-pixel $uvk$, we obtain three feature tensors: $\mathbf{P}\in{\mathbb{R}^{m\times1\times d}}$, $\mathbf{Z}\in{\mathbb{R}^{m\times 1\times f}}$ and $\mathbf{F}\in {\mathbb{R}^{m\times1\times 1}}$ as outputs of these convolutional layers respectively.

By the feature tensor $\mathbf{F}$, we construct a diagonal matrix $\mathbf{A}=diag(a_i)\in \mathbb{R}^{m\times m}$ with $a_i=\mathbf{F}(uvk,0,0)$ on the diagonal.
Then, we assign an adaptive value to each channel-pixel of a generic embedding $\mathbf{g}$ to obtain an episode-specific embedding $\mathbf{e}$, through the following linear mapping:
\begin{align}\label{eq_map}
    \mathbf{e} &= \mathbf{A}\circ \mathbf{g}.
\end{align}
Intuitively, each diagonal element $a_i$ represents an adaptive value for a generic embedding $\mathbf{g}\in\mathbb{R}^m$ at the location $(u,v)$ of the $k$-th channel. It is computed according to the distribution of generic embeddings of all instances within an episode $E=(S,Q)$ at the channel-pixel $uvk$, including support instances in $S$ and query instances in $Q$.

\subsection{Classification}
Let $\mathbf{E}_S\in {\mathbb{R}^{m\times n_s}}$ and $\mathbf{E}_Q\in {\mathbb{R}^{m\times n_q}}$ denote episode-adaptive embeddings of all instances from the support set ${S}$ and the query set ${Q}$ in an episode $E=
(S,Q)$, respectively, where $n_s=N\times K$ and $n_q=N\times T$. A classifier predicts classes of query instances in $Q$ based on $\mathbf{E}_S$ and $\mathbf{E}_Q$, as well as the classes of support instances in $S$. 

We use a prototypical network \cite{prototypical} for classification. A \emph{prototype} $\mathbf{e}^t$ is calculated for each class $t$ according to the episode-specific embeddings of all instances in $S$ of class $t$, where $\mathbf{e}_i$ stands for the episode-specific embedding of the $i$-th instance in $S$ for the class $t$.
\begin{equation}
    \mathbf{e}^t=\frac{1}{K}\sum_{i=1}^K \mathbf{e}_i
\end{equation}

 Let $d(\cdot , \cdot )$ denote a distance between two instance embeddings and  $\mathbf{e}_i$ be an episode-specific embedding of a query instance $\mathbf{x}_i$ in $Q$. Then, the probability that $\mathbf{x}_i$ belongs to a class $t$ is calculated as:
\begin{equation}
    p(y=t|\mathbf{e}_i)=\frac{\exp(-d(\mathbf{e}_i,\mathbf{e}^t))}{\sum_{j=1}^N \exp(-d(\mathbf{e}_i,\mathbf{e}^j))}
\end{equation}
The choice of $d(\cdot , \cdot )$ depends on assumptions about data distribution in the episode-specific embedding space. We use the Euclidean distance to define $d(\mathbf{e}_i,\mathbf{e}_j)=|\!|\mathbf{e}_i-\mathbf{e}_j|\!|_{2}$, where $|\!|\hspace{0.15cm}|\!|_2$ is the $l^2$ norm.
We thus predict the class $\hat{y}_i$ of $\mathbf{x}_i$ by assigning it to the same class as its nearest prototype:
\begin{equation}
    \hat{y}_i = \arg \max_{t} p(y=t|\mathbf{e}_i).  
\end{equation}
The classifier is optimized by minimizing a cross-entropy loss which averages over the losses of all query instances $\mathbf{x}_i$ in $Q$ w.r.t. their true class $y_i$:
\begin{equation}
    \mathcal{L}=-\frac{1}{n_q}\sum_{i=1}^{n_q} \log{p(y=y_i|\mathbf{e}_i)}
\end{equation}\vspace{-0.3cm}

\section{Experiments}
We evaluate our method to answer the following research questions: [\textbf{Q1.}] How does our method perform against the state-of-the-art models for few-shot classification tasks? [\textbf{Q2.}] How does our method perform against the state-of-the-art models for semi-supervised classification tasks? [\textbf{Q3.}] Is our method robust to different generic embedding networks and different classifiers? [\textbf{Q4.}] How effectively our method can leverage  instances from a query set for improving performance? We also conduct a case study to visualize how instance embeddings are changed from a general embedding space to an episode-specific embedding space.

\begin{table*}[t]
    \centering
    \resizebox{0.95\textwidth}{!}{%
   \begin{tabular}{M{0.25\textwidth} M{0.15\textwidth}   *{4}{M{0.12\textwidth}}}
    \hline
          Model & Backbone & {5-way} $1$-shot & {5-way} $5$-shot & {10-way} $1$-shot & {10-way} $5$-shot\\
          \hline\hline
          MatchingNets \cite{matchingnets} & ConvNet-4 & 43.60 & 55.30 & - & -\\
         MAML \cite{maml} & ConvNet-4 & 48.70 & 63.11 & 31.27 & 46.92 \\
         Reptile \cite{reptile} & ConvNet-4 & 47.07 & 62.74 & 31.10 & 44.66\\
PROTO \cite{prototypical} & ConvNet-4 & 46.14 & 65.77 & 32.88 & 49.29\\
RelationNet \cite{relationnets} & ConvNet-4 & 51.38 & 67.07 & 34.86 & 47.94\\
Label Propagation \cite{propagation} & ConvNet-4  & 52.31 & 68.18 & 35.23 & 51.24\\
TPN \cite{propagation} & ConvNet-4 & 53.75 & 69.43 & 36.62 & 52.32\\

GNN \cite{few_gnn} & ConvNet-4 & 50.33 & 66.41 & - & - \\
EGNN \cite{edgelabel} & ConvNet-4  & 59.18 & 76.37 & - & -\\
DPGN \cite{dpgn} & ConvNet-4 & 66.01 & 82.83 & - & -\\ 
\hline
\textbf{EA-PROTO} (ours) & \textbf{ConvNet-4} & \textbf{92.95} & \textbf{96.55} & \textbf{67.66} & \textbf{77.64}\\
\hline\hline
MetaGAN \cite{metagan} & ResNet-12 & 52.71 & 68.63 & - & -\\
TADAM \cite{tadam} & ResNet-12 & 58.50 & 76.70 & - & - \\
MetaOptNet \cite{opt} & ResNet-12 & 62.64 & 78.63 & - & -\\
FEAT \cite{set2set} & ResNet-12 & 66.79 & 82.05 & - & - \\
DPGN \cite{dpgn} & ResNet-12 & 67.77 & 84.60 & - & - \\

\hline
\textbf{EA-PROTO} (ours) & \textbf{ResNet-12} & \textbf{93.67} & \textbf{96.87} & \textbf{70.08} & \textbf{77.78}\\
\hline
    \end{tabular}%
    }
    \caption{Few-shot classification accuracies on miniImageNet.}
    \label{table_mini}\vspace{-1cm}
\end{table*}

\subsection{Datasets}
We conduct experiments on three benchmark datasets: \emph{miniImageNet}, \emph{tieredImageNet} and \emph{CIFAR-FS}. The first two datasets are subsets of ImageNet in different scales, containing RGB images of $84\times 84$ \cite{matchingnets,edgelabel}. Besides, \emph{CIFAR-FS} is a subset of CIFAR-100, containing images of $32\times 32$ \cite{cifar-fs}. 

\subsection{Experimental setup}
\vspace{-0.1cm}
\noindent\textbf{{Generic embedding networks.~}}Experiments are conducted on two widely-used backbones for generic embeddings: ConvNet-4 and ResNet-12\cite{prototypical,maml,tadam,meta-trans,opt}.
The ConvNet-4 network has four convolutional blocks. Each convolutional block begins with a $3\times 3$ $2$D convolutional layer, followed by a batch normalization (BN) layer, a $2\times 2$ max-pooling layer and a ReLU activation layer. 
The ResNet-12 network has four residual blocks with channels of $64$, $128$, $256$, and $64$. Each residual block contains three convolutional blocks, which uses a $3\times 3$ convolutional kernel, followed by a BN layer and a LeakyReLU activation layer.


\smallskip
\noindent\textbf{{Classifiers.~}}We consider two types of classifiers in experiments: prototypical network \cite{prototypical} and transductive propagation network \cite{propagation}. Thus, we have two variants of EAEN: (1) \emph{Episode Adaptive Prototypical Networks} (EA-PROTO) uses prototypical network as the classifier, and (2) \emph{Episode Adaptive Transductive Propagation Networks} (EA-TPN) uses transductive propagation network as the classifier.   

\smallskip
\noindent\textbf{{Evaluation.~}}We follow the episode training strategy for few-shot learning \cite{matchingnets,propagation}. A $N$-way $K$-shot setting is adopted for both training and testing. 
Following previous settings \cite{prototypical,propagation}, the query number is set to $15$ and the performance is measured using classification accuracy over $600$ episodes on testing data.

\smallskip
\noindent\textbf{{Parameters.~}}The init learning rate is $1e^{-3}$ for ConvNet-4 and $1e^{-4}$ for ResNet-12. In addition, the learning rate of Adam-optimizer decays by half every $10,000$ iterations.

\begin{table*}[t]
    \centering
    \resizebox{0.95\textwidth}{!}{%
    \begin{tabular}{M{0.25\textwidth} M{0.15\textwidth}   *{4}{M{0.12\textwidth}}}
    \hline
          Model &  Backbone & {5-way} $1$-shot & {5-way} $5$-shot & {10-way} $1$-shot & {10-way} $5$-shot\\
          \hline\hline
         MAML \cite{maml} & ConvNet-4 & 51.67 & 70.30 & 34.44 & 53.32 \\
         Reptile \cite{reptile} & ConvNet-4  & 48.97 & 66.47 & 33.67 & 48.04\\

PROTO \cite{prototypical} & ConvNet-4  & 48.58 & 69.57 & 37.35 & 57.839\\
IMP \cite{infinite_pro} & ConvNet-4 & 49.60 & 48.10 & - & - \\
RelationNet \cite{relationnets} & ConvNet-4  & 54.48 & 71.31 & 36.32 & 58.05\\
CovaMNET \cite{covariance} & ConvNet-4 & 51.19 & 67.65 & - & - \\

Label Propagation \cite{propagation} &  ConvNet-4  & 55.23 & 70.43 & 39.39 & 57.89\\
TPN \cite{propagation} &  ConvNet-4  & 57.53 & 72.85 & 40.93 & 59.17\\

EGNN \cite{edgelabel} &  ConvNet-4  & 63.52 & 80.24 & - & -\\
DPGN \cite{dpgn} &  ConvNet-4  & 69.43 & 85.92 & - & -\\
\hline
\textbf{EA-PROTO} (ours) & \textbf{ConvNet-4}  & \textbf{92.65} & \textbf{96.69} & \textbf{70.16} & \textbf{82.59} \\
\hline\hline
MetaOptNet \cite{opt} & ResNet-12 & 65.81 & 81.75 & - & -\\
FEAT \cite{set2set} & ResNet-12  & 70.80 & 84.79 & - & - \\

DPGN \cite{dpgn} & ResNet-12  & 72.45 & 87.24 & - & -\\
\hline
\textbf{EA-PROTO} (ours) & \textbf{ResNet-12}  & \textbf{91.56} & \textbf{97.02} & \textbf{74.50} & \textbf{83.34} \\
\hline
    \end{tabular}%
    }
    \caption{Few-shot classification accuracies on tieredImageNet.}
    \label{table_tiered}\vspace{-1cm}
\end{table*}

\subsection{Few-shot Learning}

\begin{wraptable}{r}{0.5\textwidth}
    \centering
    \vspace{-1.5cm}
    \resizebox{0.5\textwidth}{!}{%
   \begin{tabular}{M{0.25\textwidth} M{0.14\textwidth}   *{2}{M{0.08\textwidth}}}
    \hline
     \multirow{2}{*}{Model}& \multirow{2}{*}{Backbone}& {5-way} &5-way\\
      &   & $1$-shot & $5$-shot\\
     \hline\hline
    MAML$\dagger$ \cite{maml}& ConvNet-4 & 58.90 & 71.50\\
    PROTO$\dagger$ \cite{prototypical} & ConvNet-4 & 55.50 & 72.00 \\
    RelationNet$\dagger$ \cite{relationnets} & ConvNet-4 & 55.00 & 69.30 \\
    \textbf{DPGN} \cite{dpgn} & \textbf{ConvNet-4} & \textbf{76.40} & \textbf{88.40}\\\hline
    EA-PROTO (ours)& ConvNet-4 & 74.01 & 80.02\\ 
    \hline
       \end{tabular}%
       }
    \caption{Few-shot classification accuracies on CIFAR-FS, where $\dagger$ indicates that the results are from \cite{dpgn}.} 
    \label{table_cifar}\vspace{-0.6cm}
\end{wraptable}

To evaluate the effectiveness of our method for few-shot learning, we compare EA-PROTO against the state-of-the-art methods. As CIFAR-FS is a small dataset, we follow \cite{cifar-fs,dpgn} to consider $5$-way $1$-shot and $5$-way $5$-shot. The results are shown in Tables \ref{table_mini}-\ref{table_cifar}.

From Tables \ref{table_mini}-\ref{table_tiered}, we see that EA-PROTO significantly outperform all baselines on both miniImageNet and tieredImageNet, regardless of using ConvNet-4 or ResNet-12 as the generic embedding network. Specifically, 1) on miniImageNet, EA-PROTO improves upon the best results of the baselines by a margin $25.9\%$ in $5$-way $1$-shot and $12.27\%$ in $5$-way $5$-shot; 2) on tieredImageNet, EA-PROTO improves upon the best results of the baselines by a margin $19.11\%$ in $5$-way $1$-shot and $9.78\%$ in $5$-way $5$-shot.

Table \ref{table_cifar} shows that EA-PROTO performs better than all other models except DPGN. In $5$-way $1$-shot, EA-PROTO improves about $16\%$ on average than the other three models, but performs $2\%$ slightly worse than DPGN. 
The reason why DPGN has a better performance than EA-PROTO is that the low resolution images ($32\times 32$) in CIFAR-FS make generic embeddings of instances contain less useful information compared with those from miniImageNet and tieredImageNet ($84\times 84$). 
This limits the expressiveness of episode specific embeddings learned from CIFAR-FS and accordingly hinders the performance of EA-PROTO. 
DPGN concatenates the output of the last two layers of a generic embedding network as generic embeddings. Hence, DPGN performs better than all the other models on CIFAR-FS. 

\vspace{-0.2cm} 
\begin{table*}[h]
    \centering
    \resizebox{0.95\textwidth}{!}{%
    \begin{tabular}{M{0.3\textwidth} M{0.2\textwidth} *{5}{M{0.08\textwidth}}}
    \hline

    \multirow{ 2}{*}{Model}& \multirow{ 2}{*}{Training Strategy} & \multicolumn{5}{c}{Labeled Ratio (5-way 5-shot)}\\
    \cline{3-7}
     &  & 20\% & 40\% & 60\% & 80\% & 100\% \\
    \hline\hline
    GNN \cite{few_gnn} & Supervised & 50.33 & 56.91  & - & -&  66.41\\
    GNN-Semi \cite{few_gnn} & Semi-supervised & 52.45 & 58.76  & - & -& 66.41\\

    EGNN \cite{edgelabel} & Supervised & 59.18 & -& - & - & 76.37\\
    EGNN-Semi \cite{edgelabel} & Semi-supervised & 63.62 & 64.32 & 66.37 & - & 76.37\\
    \hline
    \textbf{EA-PROTO} (ours) & \textbf{Supervised} & \textbf{92.95} & \textbf{95.03} & \textbf{95.89} & \textbf{96.24} & \textbf{96.55}\\
    \textbf{EA-PROTO-Semi} (ours) & \textbf{Semi-supervised} & \textbf{93.01} & \textbf{95.14} & \textbf{96.05} & \textbf{96.43} & \textbf{96.55}\\
    \hline
    \end{tabular}%
    }
    \caption{Semi-supervised classification accuracies
on miniImageNet. X-Semi stands for a model X which uses unlabeled instances in a support set. While X stands for a model that only use labeled instances in a support set.}
    \label{table_semi}\vspace{-1.5cm}
\end{table*}

\begin{figure*}[t]
    \centering
    \includegraphics[width=0.9\textwidth]{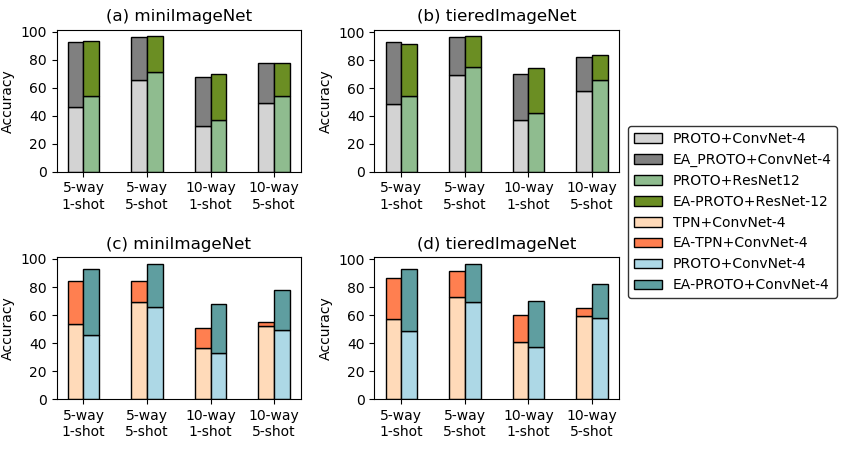}\vspace{-0.2cm}
   \caption{Classification accuracies with classifiers and generic embedding networks.}
    \label{fig_robust}\vspace{-0.5cm}
\end{figure*}

\subsection{Semi-supervised Learning}
For semi-supervised learning, we conduct experiments on miniImageNet in the $5$-way $5$-shot setting. Following \cite{few_gnn,edgelabel}, we partially label the same number of instances for each class in a support set, and consider two training strategies: (1) \emph{supervised} -- training with only labeled instances in a support set; (2) \emph{semi-supervised} -- training with all instances in a support set. These two strategies only differ in whether or not they use unlabeled instances in a support set. 

 The results are shown in Table \ref{table_semi}. We find that: 1) Semi-supervised models achieve better performance compared with their corresponding supervised models. This is because unlabeled instances in a support set help in classification. 2) EA-PROTO-Semi consistently achieves the best performance under all different labeled ratios \{20\%, 40\%, 60\%, 80\%, 100\%\}. EA-PROTO-Semi outperforms EGNN-Semi and GNN-Semi significantly. The margin between EA-PROTO-Semi and EGNN-Semi is about $30\%$ when the labeled ratio is $20\%$, and decreases to $20\%$ when the labeled ratio is $100\%$. 3) EA-PROTO-Semi has a smaller performance gap between the labeled ratios from $20\%$ to $100\%$ than the other models. This is due to the fact that episode specific embeddings in EA-PROTO are learned from all instances in an episode, regardless whether they are labeled or not, while the other models rely only on labeled instances.

\begin{table*}[t]
    \centering
    
    \resizebox{0.95\textwidth}{!}{%
    \begin{tabular}{M{0.2\textwidth} M{0.13\textwidth} M{0.14\textwidth}  *{4}{M{0.1\textwidth}}}
    \hline

          Model & Backbone &  Dataset & 5-way $1$-shot & 5-way $5$-shot & 10-way $1$-shot & 10-way $5$-shot\\
          \hline
\hline
TPN \cite{propagation} & ConvNet-4 &  & 53.75 & 69.43 & 36.62 & 52.32\\
EA-TPN-S & ConvNet-4 & miniImageNet & 50.30 & 68.41 & 36.15 & 52.11\\
EA-TPN & ConvNet-4 &  & 84.01 & 84.43 & 50.73 & 54.85 \\
\hline
PROTO \cite{prototypical} & ConvNet-4 &  & 46.14 & 65.77 & 32.88 & 49.29\\
EA-PROTO-S & ConvNet-4 & miniImageNet & 49.64 & 67.42 & 34.08 & 48.94 \\
EA-PROTO & ConvNet-4 &  & 92.95 & 96.55 & 68.08 & 78.99\\
\hline
    \end{tabular}%
    }
    \caption{Results for an ablation study, where EA-PROTO-S and EA-TPN-S refer to a variant of the methods EA-PROTO and EA-TPN, respectively, which use only instances in a support set to learn their episode-specific embeddings.}
    \label{table_compare}\vspace{-0.8cm}
\end{table*}


\vspace{-0.2cm}
\subsection{Robustness Analysis} 
To evaluate the robustness of our method, we conduct experiments under different combinations of generic embedding networks and classifiers. 
The results on miniImageNet and tieredImageNet are presented in Figure \ref{fig_robust}. 

We observe that: (1) Our method is robust to different generic embedding networks. We compare performance of PROTO and EA-PROTO when using ConvNet-4 and ResNet-12 as the generic embedding network separately on miniImageNet and tieredImageNet. Figure \ref{fig_robust}(a)-(b) shows that our method consistently yields improvement, no matter which generic embedding network or dataset is used.
(2) Our method is robust to different classifiers. We compare the performance of PROTO against EA-PROTO, as well as TPN against EA-TPN, when using convNet-4 as the generic embedding network. In Figure \ref{fig_robust}(c)-(d), both EA-PROTO and EA-TPN perform better than PROTO and TPN, respectively, on both miniImageNet and tieredImageNet datasets.


\vspace{-0.2cm}
\subsection{Ablation Analysis}
To study how effectively our method can use instances from a query set for improving performance, we conduct an ablation analysis that compares EA-PROTO and EA-TPN (using instances from both support and query sets) against EA-PROTO-S and EA-TPN-S (using only instances in a support set).
The results are shown in Table \ref{table_compare}. 

We observe that: 1) A large performance gap exists between EA-PROTO and EA-PROTO-S, and similarly between EA-TPN and EA-TPN-S. This is due to the fact that there are more instances in a query set than instances in a support set. In $5$-way $1$-shot setting, the size of a query set is $75$ while the size of a support set is $5$. Thus, by utilizing $80$ instances from both support and query sets, EA-PROTO and EA-TPN can generate better episode specific embeddings than EA-PROTO-S and EA-TPN-S which only use $5$ instances from a support set. 
2) EA-PROTO-S performs slightly better than PROTO, whereas EA-TPN-S performs slightly worse than TPN. This is because episode adaptive embeddings cannot be effectively computed from instances of a support set. When the number of instances in a support set is limited, computing episode adaptive embeddings only from instances of a support set may even harm performance.

\begin{wrapfigure}{r}{0.5\textwidth}
\vspace{-0.5cm}
\centering
\includegraphics[width=0.5\textwidth]{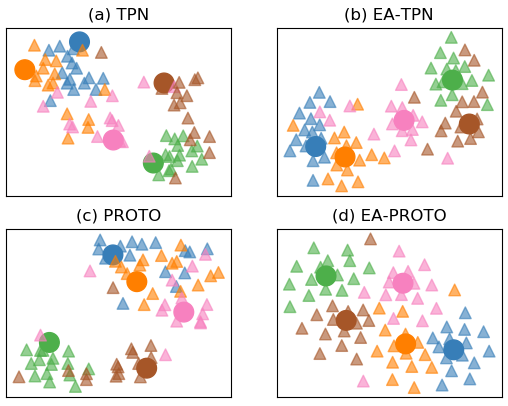}
\caption{t-SNE for image embeddings on miniImageNet under the $5$-way $1$-shot setting. Circles and triangles in each subfigure stand for image embeddings in the support and query sets of an episode, respectively. Different colors indicate different classes.}
\label{fig_case}\vspace{-0.6cm}
\end{wrapfigure} 

\subsection{Case Study}
To explore how effectively our method maps instances into an episode-specific embedding space, we conduct a case study using images from miniImageNet dataset. We compare generic embeddings learned from PROTO and TPN with episode-specific embeddings learned from EA-PROTO and EA-TPN in the $5$-way $1$-shot setting, where ConvNet-4 is used as the generic embedding network. We use t-SNE \footnote{https://lvdmaaten.github.io/tsne/} to visualize embeddings.

Figure \ref{fig_case}(a)-(b) shows the t-SNE maps of image embeddings in an episode being produced by TPN and EA-TPN, respectively, while Figure \ref{fig_case}(c)-(d) shows the t-SNE maps of image embeddings in an episode being produced by PROTO and EA-PROTO, respectively. Each triangle represents an image embedding in a query set, and each circle represents an image embeddng in a support set. 
From Figure \ref{fig_case}(a)-(b), we can see that the location of each circle is closer to the center of triangles for the same class in Figure \ref{fig_case}(b) than in Figure \ref{fig_case}(a). It indicates that the instance embeddings learnt by EA-TPN provide more discriminative information for classifying these instances accurately than the instance embeddings learnt by TPN. There is a similar trend in Figure \ref{fig_case}(c)-(d), indicating EA-PROTO captures more discriminative features (i.e., episode-specific features) in its embeddings than PROTO for classification. These show that episode adaptive embeddings are more discriminative than generic embeddings, which helps improve classification. 


\section{Conclusion}
In this work, we have proposed EAEN \footnote{Our code is available at https://www.dropbox.com/s/cll23kem3yswg96/EAEN.zip?dl=0}, a novel approach for learning episode-specific instance embeddings in few-shot learning, EAEN maps generic embeddings to episode-specific embeddings using an episode adaptive module which is learnt from the probability distribution of generic embeddings at each channel-pixel of all instances within an episode. Such episode-specific embeddings are discriminative, and can thus help classify instances in episodes, even when only a few labelled instances are available. Our experimental results on three benchmark datasets have empirically verified the effectiveness and robustness of EAEN. It is shown that EAEN significantly improves classification accuracy compared with the-state-of-the-art methods.

%
%
%
%


\begin{thebibliography}{10}
\providecommand{\url}[1]{\texttt{#1}}
\providecommand{\urlprefix}{URL }
\providecommand{\doi}[1]{https://doi.org/#1}

\bibitem{infinite_pro}
Allen, K.R., Shelhamer, E., Shin, H., Tenenbaum, J.B.: Infinite mixture
  prototypes for few-shot learning. arXiv preprint arXiv:1902.04552  (2019)

\bibitem{train_maml}
Antoniou, A., Edwards, H., Storkey, A.: How to train your maml. In: ICLR (2018)

\bibitem{cifar-fs}
Bertinetto, L., Henriques, J.F., Torr, P.H., Vedaldi, A.: Meta-learning with
  differentiable closed-form solvers. arXiv preprint arXiv:1805.08136  (2018)

\bibitem{baseline++}
Chen, W.Y., Liu, Y.C., Kira, Z., Wang, Y.C.F., Huang, J.B.: A closer look at
  few-shot classification. arXiv preprint arXiv:1904.04232  (2019)

\bibitem{maml}
Finn, C., Abbeel, P., Levine, S.: Model-agnostic meta-learning for fast
  adaptation of deep networks. In: ICML. pp. 1126--1135 (2017)

\bibitem{few_gnn}
Garcia, V., Estrach, J.B.: Few-shot learning with graph neural networks. In:
  ICLR (2018)

\bibitem{han2018meta}
Han, C., Shan, S., Kan, M., Wu, S., Chen, X.: Meta-learning with individualized
  feature space for few-shot classification  (2018)

\bibitem{augment2}
Hariharan, B., Girshick, R.: Low-shot visual recognition by shrinking and
  hallucinating features. In: ICCV. pp. 3018--3027 (2017)

\bibitem{edgelabel}
Kim, J., Kim, T., Kim, S., Yoo, C.D.: Edge-labeling graph neural network for
  few-shot learning. In: CVPR. pp. 11--20 (2019)

\bibitem{opt}
Lee, K., Maji, S., Ravichandran, A., Soatto, S.: Meta-learning with
  differentiable convex optimization. In: CVPR. pp. 10657--10665 (2019)

\bibitem{task_related}
Li, H., Eigen, D., Dodge, S., Zeiler, M., Wang, X.: Finding task-relevant
  features for few-shot learning by category traversal. In: CVPR. pp. 1--10
  (2019)

\bibitem{covariance}
Li, W., Xu, J., Huo, J., Wang, L., Gao, Y., Luo, J.: Distribution consistency
  based covariance metric networks for few-shot learning. In: AAAI. vol.~33,
  pp. 8642--8649 (2019)

\bibitem{propagation}
Liu, Y., Lee, J., Park, M., Kim, S., Yang, E., Hwang, S.J., Yang, Y.: Learning
  to propagate labels: Transductive propagation network for few-shot learning.
  In: ICLR. International Conference on Learning Representations, ICLR (2019)

\bibitem{reptile}
Nichol, A., Achiam, J., Schulman, J.: On first-order meta-learning algorithms.
  arXiv preprint arXiv:1803.02999  (2018)

\bibitem{tadam}
Oreshkin, B., L{\'o}pez, P.R., Lacoste, A.: Tadam: Task dependent adaptive
  metric for improved few-shot learning. In: NeurIPS. pp. 721--731 (2018)

\bibitem{prototypical}
Snell, J., Swersky, K., Zemel, R.: Prototypical networks for few-shot learning.
  In: NeurIPS. pp. 4077--4087 (2017)

\bibitem{meta-trans}
Sun, Q., Liu, Y., Chua, T.S., Schiele, B.: Meta-transfer learning for few-shot
  learning. In: CVPR. pp. 403--412 (2019)

\bibitem{relationnets}
Sung, F., Yang, Y., Zhang, L., Xiang, T., Torr, P.H., Hospedales, T.M.:
  Learning to compare: Relation network for few-shot learning. In: CVPR. pp.
  1199--1208 (2018)

\bibitem{matchingnets}
Vinyals, O., Blundell, C., Lillicrap, T., Wierstra, D., et~al.: Matching
  networks for one shot learning. In: NeurIPS. pp. 3630--3638 (2016)

\bibitem{task_aware}
Wang, X., Yu, F., Wang, R., Darrell, T., Gonzalez, J.E.: Tafe-net: Task-aware
  feature embeddings for low shot learning. In: CVPR. pp. 1831--1840 (2019)

\bibitem{augment1}
Wang, Y.X., Girshick, R., Hebert, M., Hariharan, B.: Low-shot learning from
  imaginary data. In: CVPR. pp. 7278--7286 (2018)

\bibitem{fine-grain}
Wei, X.S., Wang, P., Liu, L., Shen, C., Wu, J.: Piecewise classifier mappings:
  Learning fine-grained learners for novel categories with few examples. TIP
  \textbf{28}(12),  6116--6125 (2019)

\bibitem{dpgn}
Yang, L., Li, L., Zhang, Z., Zhou, X., Zhou, E., Liu, Y.: Dpgn: Distribution
  propagation graph network for few-shot learning. In: CVPR. pp. 13390--13399
  (2020)

\bibitem{set2set}
Ye, H.J., Hu, H., Zhan, D.C., Sha, F.: Few-shot learning via embedding
  adaptation with set-to-set functions. In: CVPR. pp. 8808--8817 (2020)

\bibitem{metagan}
Zhang, R., Che, T., Ghahramani, Z., Bengio, Y., Song, Y.: Metagan: An
  adversarial approach to few-shot learning. In: NeurIPS. pp. 2365--2374 (2018)

\end{thebibliography}


\end{document}